\documentclass[lettersize,journal]{IEEEtran}
\usepackage{amsmath,amsfonts}
\usepackage{algorithmic}
\usepackage{array}
\usepackage[caption=false,font=normalsize,labelfont=sf,textfont=sf]{subfig}
\usepackage{textcomp}
\usepackage{stfloats}
\usepackage{url}
\usepackage{verbatim}
\usepackage{graphicx}
\usepackage{gensymb}
\usepackage{tikz}
\usepackage{subfig}
\usepackage[table]{xcolor}
\usetikzlibrary{arrows.meta, positioning}
\usepackage{cite}
\usepackage{float}
\usepackage{comment}
\usepackage[ruled,vlined]{algorithm2e}
\begin{document}

\title{Shared Voxel-Map-Based Cooperative Indoor UAV Guidance with a Multi-Agent Soft Actor-Critic Controller}

\author{\IEEEauthorblockN{Thomas Hickling, Dylan Wynne, Yu Su and Nabil Aouf}\\
\IEEEauthorblockA{School of Science and Technology\\
City St George's University of London\\
London, United Kingdom\\
Email: Tom.Hickling, Dylan.Wynne, Yu.Su, Nabil.Aouf@city.ac.uk}
\thanks{}
\thanks{}}

\maketitle

\begin{abstract}
This paper presents a cooperative indoor UAV guidance framework that combines a shared voxel-map world model with a multi-agent Soft Actor-Critic (MASAC) controller. Multiple drones fuse 360$^\circ$ LiDAR observations into a common world-frame occupancy map, which is converted into a compact bird's-eye-view (BEV) representation and provided to each agent as an ego-aligned local crop. This \emph{integrate-in-world, act-in-ego} design enables consistent multi-UAV spatial fusion whilst retaining decentralised continuous control. The policy combines BEV map features, near-field obstacle observations, and compact goal and peer-state information within a centralised-training, decentralised-execution framework. In simulation, the learned controller achieves a 90.3\% success rate in corridor navigation, outperforming A* planning, an artificial potential field controller, and a prior guidance method. To address residual sim-to-real mismatch, the simulation-trained policy is further adapted using offline imitation fine-tuning from real-world data. Real-world experiments in GNSS-denied indoor environments demonstrate stable two-UAV cooperative operation across increasingly challenging obstacle layouts. The results show that shared voxel-map representations provide an effective and scalable spatial substrate for learned cooperative indoor UAV guidance.

\textit{Note to Practitioners}---Deploying multiple UAVs indoors is difficult because each drone must navigate around obstacles, avoid other vehicles, and operate without GNSS, all while relying on limited onboard sensing and communication. This paper presents a cooperative guidance approach for that setting. Each drone contributes LiDAR data to a shared map, allowing the team to make better navigation decisions than would be possible from local sensing alone. The approach outperformed conventional planning and reactive avoidance methods in simulation and was successfully deployed on two real UAVs in a GNSS-denied indoor arena. A key practical finding is that successful deployment depends not only on the AI controller, but also on the surrounding autonomy stack, including localisation, communication, map fusion, and safety monitoring. Another important lesson is that simulation training alone was not sufficient: some real-world adaptation was required before reliable hardware operation was achieved. The current study is limited to two drones and indoor environments, but the same ideas could be useful for inspection, warehouse automation, search and rescue, and other multi-robot systems that need shared situational awareness in confined spaces.
\end{abstract}


\section{Introduction}

\subsection{Research background}

Autonomous indoor flight in GNSS-denied environments remains a challenging problem for unmanned aerial vehicles (UAVs), particularly when multiple platforms must operate simultaneously in confined and cluttered spaces. In such settings, reliable collision avoidance must be maintained despite partial observability, occlusions, sensing noise, and localisation uncertainty, whilst sufficient spatial context is needed to negotiate longer-horizon geometric structures such as corridors, junctions, and dead ends. In the multi-UAV case, the problem is further complicated by dynamic inter-agent interactions, communication constraints, and the requirement to safely share limited airspace.

Classical navigation pipelines typically decompose the problem into mapping, planning, and control, for example by combining occupancy grids or voxel maps with graph-based planners such as A* and a local tracking controller \cite{hart1968formal,moravec1985high,thrun2002probabilistic,hornung2013octomap}. Such methods remain attractive because they are interpretable and geometrically grounded, but their practical performance can degrade when mapping is noisy, map updates are delayed, replanning is intermittent, or neighbouring UAVs are handled only as static obstacles. At the opposite end of the spectrum, reactive approaches such as artificial potential fields (APF) are computationally efficient and straightforward to deploy, but are prone to local minima and dead-end failures in cluttered environments when decisions are driven primarily by local attractive and repulsive cues \cite{khatib1986real}.

Deep reinforcement learning (DRL) provides an alternative by learning continuous control policies directly from sensory observations. In particular, entropy-regularised off-policy methods such as Soft Actor-Critic (SAC) are attractive for robotics because they offer strong performance in continuous action spaces together with improved optimisation stability and sample efficiency \cite{haarnoja2018soft,haarnoja2018soft2}. In multi-agent settings, centralised training with decentralised execution (CTDE) is widely adopted to mitigate non-stationarity during learning by allowing richer training-time value estimation whilst preserving local execution policies at deployment \cite{lowe2017multi}.

Therefore, there remains a need for navigation frameworks that can effectively integrate global spatial structure with local decision-making, while maintaining scalability and robustness in multi-UAV indoor environments.

\subsection{Previous work}

\subsubsection{Structured mapping, planning, and reactive guidance}

Occupancy-grid mapping remains a core approach in robotic navigation, enabling noisy range measurements to be fused into a consistent spatial representation \cite{moravec1985high,thrun2002probabilistic}. For 3D environments, OctoMap provides an efficient probabilistic octree-based occupancy framework that has been widely adopted in robotic mapping and planning \cite{hornung2013octomap}. In indoor UAV navigation, such structured map representations are attractive because they decouple perception from control and provide an interpretable substrate for planning, debugging, and safety analysis. However, their effectiveness depends strongly on update stability, latency, and the ability to incorporate information from multiple robots without excessive communication overhead.

Graph-based planners such as A* remain widely used because of their completeness and optimality properties under suitable modelling assumptions \cite{hart1968formal}. Reactive methods such as artificial potential fields (APF) offer lower computational cost and higher update rates, but are prone to local minima and oscillatory behaviour in cluttered or dead-end environments \cite{khatib1986real}. These trade-offs motivate hybrid approaches that retain explicit geometric structure for mid-field spatial reasoning whilst using learned control for robust local decision-making under imperfect sensing and dynamic interactions.

To balance spatial expressiveness and computational efficiency, bird's-eye-view (BEV) representations have been widely adopted, providing a compact projection of 3D environments in which relevant obstacle structures are well captured in near-horizontal slices. Beyond binary occupancy, additional channels can further enhance spatial representation; in particular, Euclidean distance transforms encode obstacle clearance and provide smooth spatial cues for planning and control \cite{felzenszwalb2012distance}. 

In multi-robot systems, cooperative mapping has long been studied as a means to improve coverage and reduce uncertainty through shared spatial information \cite{burgard2005coordinated}. However, in decentralised settings, communication bandwidth and latency remain major constraints, motivating the development of compressed representations and selective information-sharing strategies \cite{bayer2026decentralized,zhang2024reducing}.

\subsubsection{Reinforcement learning, CTDE, and sim-to-real adaptation}

SAC has become a widely adopted method for continuous control due to its off-policy formulation, entropy regularisation, and strong empirical stability \cite{haarnoja2018soft,haarnoja2018soft2}. In navigation problems, dense reward shaping is often employed to improve learning efficiency, with potential-based shaping providing a principled framework for introducing informative progress signals \cite{ng2003shaping}. In multi-agent settings, centralised training with decentralised execution (CTDE) is commonly used to mitigate non-stationarity, allowing training-time critics to leverage richer joint information while maintaining decentralised policies at deployment \cite{lowe2017multi,su2021value}.

For UAV guidance, DRL has been applied to tasks such as obstacle avoidance, path planning, and cooperative motion generation. Existing approaches range from reactive policies driven by local sensing to hybrid methods that integrate learned policies with planning structures or imitation signals. While reactive policies can achieve high-frequency collision avoidance, they often struggle in geometrically constrained environments where mid-field spatial context is required. Multi-agent formulations further introduce challenges related to peer interaction, partial observability, and safety constraints, and many existing methods simplify the problem by assuming access to peer states or to a globally consistent map.

Bridging the sim-to-real gap remains a major challenge in learned UAV control due to discrepancies in sensing, dynamics, and environmental variability. To address this issue, recent work has explored combining reinforcement learning with demonstrations, corrective supervision, or additional fine-tuning stages to improve transfer robustness \cite{xing2024bootstrapping,joshi2024sim,jiang2025transic,torne2024reconciling,cheng2025generalizable,yin2025rapidly,ankile2025residual}. These studies highlight the effectiveness of incorporating limited real-world data after simulation training, although challenges remain in achieving reliable generalisation in complex and cooperative navigation scenarios.

\subsection{Present work}

Existing approaches to UAV navigation typically either rely on structured geometric representations for planning or leverage learning-based policies for control. Whilst mapping-based methods provide interpretability and global consistency, they often lack adaptability in dynamic and uncertain environments. In contrast, learning-based approaches offer greater flexibility but are frequently limited by partial observability and insufficient exploitation of spatial structure.

To address these limitations, this work adopts a hybrid perspective that combines the structural advantages of explicit geometric mapping with the adaptability of learned control. Rather than learning directly from raw observations, the proposed framework enables multiple UAVs to fuse 360$^\circ$ LiDAR measurements into a shared world-frame voxel occupancy map, which is subsequently transformed into a compact BEV representation for policy input.

A key design principle is the separation between \emph{mapping} and \emph{control frames}. Spatial information is integrated in a fixed world frame to ensure consistent multi-UAV fusion, while each agent operates on an ego-centric BEV crop aligned with its body frame. This \emph{integrate-in-world, act-in-ego} formulation reduces the learning burden associated with rotational and translational invariance while preserving a shared and scalable spatial representation.

The novelty of this work does not lie in individual components such as voxel maps, BEV representations, or SAC, but in their integration into a unified cooperative indoor guidance framework. The shared voxel-derived BEV representation serves as a common spatial substrate for decentralised continuous control, enabling multiple UAVs to exploit fused mid-field spatial context without increasing the dimensionality of policy inputs with team size. This results in a system that remains geometrically interpretable, scalable to larger teams, and capable of end-to-end continuous cooperative navigation in cluttered indoor environments.

In contrast to prior approaches that primarily utilise shared maps for planning or exploration, the proposed framework employs a shared voxel-derived BEV representation as a unified perception-to-control interface for decentralised learned guidance. Each UAV contributes sensed geometry to a common world-frame representation, whilst each agent consumes only a fixed-size ego-aligned BEV crop together with compact relative state information. This design preserves decentralised execution and prevents the policy input dimensionality from increasing with team size.

Building upon our prior work on robust DRL-based UAV guidance and cooperative decision-making in real-world settings \cite{hickling2023robust,hickling2025deep}, the present study extends this line of research to cooperative indoor navigation. The proposed framework is validated in both simulation and GNSS-denied real-world flight experiments, demonstrating its effectiveness in complex indoor environments.

To further address residual sim-to-real discrepancies encountered during deployment, an offline adaptation stage is introduced, in which the simulation-trained policy is fine-tuned through behaviour cloning using actions generated by an A* reference controller \cite{jiang2025transic}. This improves transfer robustness whilst preserving the advantages of the learned policy.

\subsection{Contributions}
The principal contributions of this paper are as follows:
\begin{itemize}
    \item A cooperative indoor UAV guidance architecture in which multiple drones fuse LiDAR observations into a shared world-frame voxel map, whilst each agent receives a fixed-size ego-aligned BEV crop for decentralised continuous control.
    
    \item A compact BEV state representation that augments binary occupancy information with a Euclidean-distance-based clearance channel, thereby providing both obstacle location and a smooth traversability surrogate for policy learning.
    
    \item A centralised-training, decentralised-execution multi-agent Soft Actor-Critic controller with decentralised actors and centralised twin critics, together with practical training measures to improve stability during large-scale experience collection and optimisation.
    
    \item A simulation study in corridor environments of increasing geometric difficulty, demonstrating that the learned cooperative guidance policy outperforms classical baselines and a prior learned guidance method.
    
    \item A real-world deployment of the cooperative guidance framework on two UAVs in a GNSS-denied indoor environment, including onboard perception, shared-map communication, safety-gated execution, and offline adaptation for sim-to-real transfer.
    
    \item An offline real-world adaptation procedure based on imitation fine-tuning from an A* reference controller, enabling incremental correction of sim-to-real mismatch and checkpoint-based selection of deployable policies.

\end{itemize}



\section{Methodology}

\subsection{Perception and shared map representation}
The perception pipeline converts raw 360\degree{} LiDAR measurements into compact control-oriented representations at two spatial scales. Near-field obstacle avoidance is supported by a lightweight angular depth feature, whereas mid-field spatial context and multi-agent information sharing are provided through a persistent shared world-frame occupancy map.

All mapping is performed in a common world frame using the NED convention. This is necessary for fusing observations from multiple UAVs into a single spatial representation. Immediately before policy inference, a fixed local window is cropped around each agent and rotated into the vehicle body-aligned frame, yielding an \emph{integrate-in-world, act-in-ego} representation. This preserves stable multi-UAV fusion whilst presenting the policy with a task-aligned local view.

\begin{figure}[!htbp]
    \centering
    \includegraphics[width=1.0\linewidth]{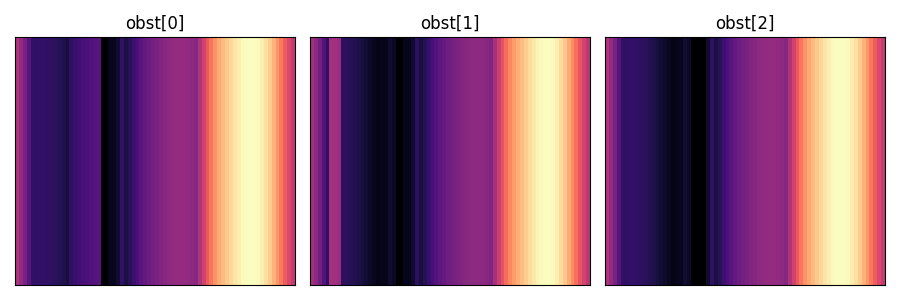}
    \caption{The three inverse depth image bands for low (left), medium (centre), and high (right)}
    \label{fig:obst_depth}
\end{figure}

The LiDAR point cloud is compressed into a fixed-size angular feature tensor represented as inverse depth in $[0,1]$, where 1 denotes near obstacles and 0 denotes distant returns. To preserve coarse vertical structure without reconstructing full 3D geometry, three elevation bands (below, level, and above) are used where appropriate as shown in Fig \ref{fig:obst_depth}. To reduce computational cost, dense scans are sub-sampled using fixed strides.

\begin{figure}[!htbp]
    \centering
    \includegraphics[width=1.0\linewidth]{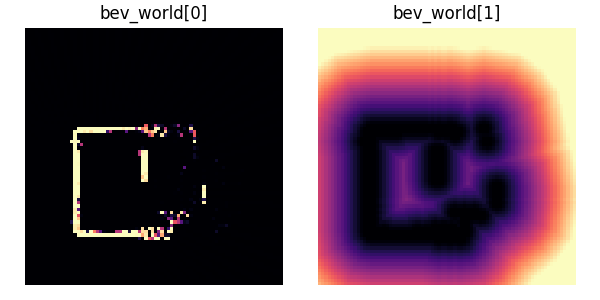}
    \caption{The shared world BEV map used cooperatively between the two drones}
    \label{fig:bev_world}
\end{figure}

\begin{figure}[!htbp]
    \centering
    \includegraphics[width=1.0\linewidth]{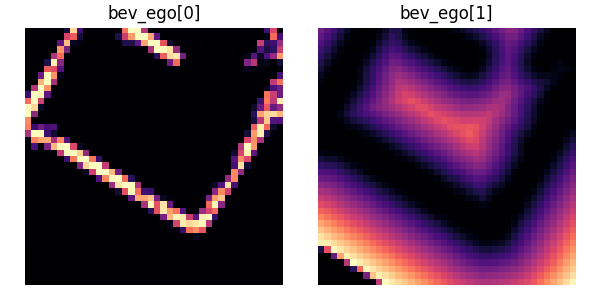}
    \caption{The ego view fed to the neural network, here right is forward of the drone}
    \label{fig:bev_ego}
\end{figure}

The shared world model is maintained as a 3D occupancy grid with resolution $0.25$\,m per cell. Free space is updated by ray-carving along each measurement ray and occupied space is incremented at the terminal hit cell. For policy input, the 3D grid is projected into a 2D BEV representation by accumulating occupancy within a vertical slice of $\pm0.5$\,m about the vehicle height. A shared world BEV map between the two drones can be found in Fig \ref{fig:bev_world}. 

A Euclidean distance transform over free cells then provides a clearance channel, clipped to a practical range for numerical stability \cite{felzenszwalb2012distance}. Each policy therefore receives a two-channel ego-aligned BEV tensor,
\begin{equation}
\{\mathrm{occupancy}, \mathrm{clearance}\}
\end{equation}
cropped to $39\times39$ cells, corresponding to $9.75\times 9.75$\,m at the chosen resolution. The ego view fed to the neural network is shown in Fig \ref{fig:bev_ego}.

In addition to the map input, each agent receives a compact goal-and-peer state vector:
\begin{equation}
\mathbf{g}=\big[d,\ \sin\theta,\ \cos\theta,\ \Delta z,\ d_{\text{peer}}\big]
\end{equation}
where $d$ is planar goal distance, $\theta$ is the ego-frame bearing to the goal, $\Delta z$ is vertical error, and $d_{\text{peer}}$ is inter-UAV separation. This provides minimal global context without increasing the dimensionality of the convolutional inputs.

\begin{figure}[!htbp]
\centering
\begin{tikzpicture}[
  node distance=0.6cm and 1.1cm,
  every node/.style={font=\footnotesize},
  box/.style={draw, rounded corners, thick, minimum width=2cm, minimum height=0.7cm, align=center},
  smallbox/.style={draw, rounded corners, thick, minimum width=2.7cm, minimum height=0.6cm, align=center},
  arr/.style={-Latex, thick}
]

\node[box] (actor_label) {\bfseries Actor $\pi_\phi$ (per agent)};

\node[box, below=0.4cm of actor_label] (inputs) {Goal/Peer};

\node[box, left=0.8cm of inputs] (obst) {Obstacle\\Features};

\node[box, right=0.8cm of inputs] (ego) {ego BEV};

\node[box, below=0.4cm of inputs] (branches) {MLP$_{\text{goal}}$};

\node[box, left=0.8cm of branches] (cnn_obst) {CNN$_{\text{obst}}$};

\node[box, right=0.8cm of branches] (cnn_ego) {CNN$_{\text{BEV}}$};

\node[box, below=0.4cm of branches] (fusion) {Concatenate \& fusion MLP};

\node[smallbox, below=0.4cm of fusion] (gauss) {Gaussian head\\$\mu(z), \log\sigma(z)$};

\node[box, below=0.4cm of gauss] (act_out) {Tanh activation\\Action $(v_x, \dot{\psi}, v_z)$};

\draw[arr] (inputs) -- (branches);
\draw[arr] (obst) -- (cnn_obst);
\draw[arr] (ego) -- (cnn_ego);
\draw[arr] (branches) -- (fusion);
\draw[arr] (cnn_obst.south east) -- (fusion.north west);
\draw[arr] (cnn_ego.south west) -- (fusion.north east);
\draw[arr] (fusion) -- (gauss);
\draw[arr] (gauss) -- (act_out);


\node[box, below=1.0cm of act_out] (critic_label) {\bfseries Centralised critics};

\node[box, below=0.4cm of critic_label] (z_stack) {Per-agent latents $\{\mathbf{z}_i\}$\\\& joint actions $\{\mathbf{a}_i\}$};

\node[box, below=0.4cm of z_stack] (crit_trunk) {Shared critic trunk\\(MLP)};

\node[smallbox, below=0.4cm of crit_trunk] (q_heads) {Twin heads\\$Q_{\theta_1}(s,a),\,Q_{\theta_2}(s,a)$};

\draw[arr] (z_stack) -- (crit_trunk);
\draw[arr] (crit_trunk) -- (q_heads);


\draw[arr, dashed] (fusion.west) -- ++(-1.0,0) |- ++(0,-5.4) node[left,pos=0.85]{latents $\mathbf{z}_i$} -- (z_stack.west);
\draw[arr, dashed] (act_out.east) -- ++(1.0,0) |- ++(0,-3.0) node[right,pos=0.85]{actions $\mathbf{a}_i$} -- (z_stack.east);

\end{tikzpicture}
\caption{The actor-critic architecture. The actor fuses obstacle features, goal/peer encoding, and ego-centric BEV into a latent representation and a Gaussian action head. A centralised critic receives stacked per-agent latents and joint actions and outputs twin value estimates for SAC.}
\label{fig:masac-architecture-compact}
\end{figure}
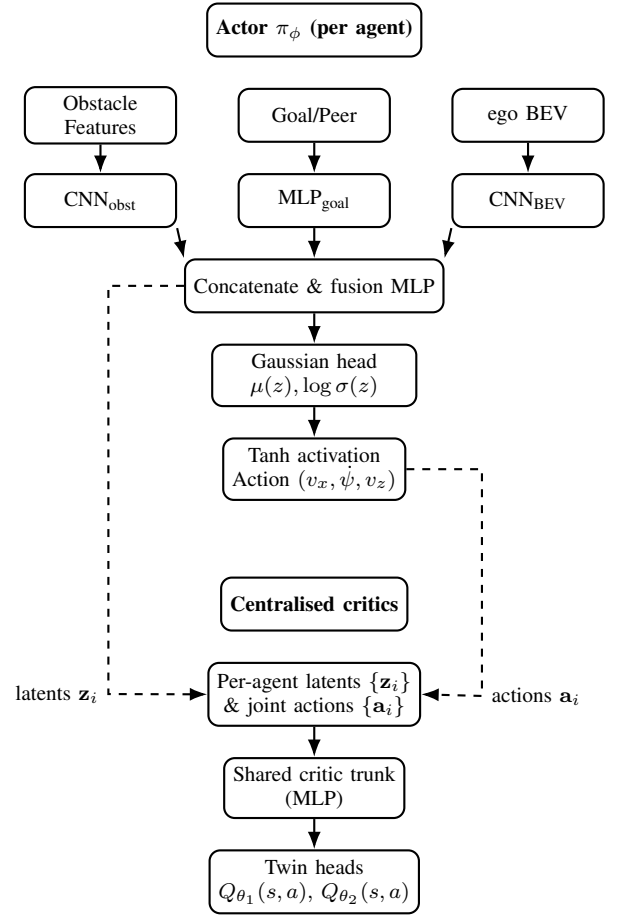

\subsection{MASAC policy and critic architecture}

To enable stable learning in continuous control with multi-agent interactions, we adopt a multi-agent extension of Soft Actor-Critic (MASAC), which combines entropy-regularised policy optimisation with CTDE. A shared multi-branch actor is used for all agents within the MASAC framework. The actor-critic architecture is shown in Fig \ref{fig:masac-architecture-compact}. The actor processes three input streams: i) angular near-field obstacle features, ii) the ego-aligned BEV map, and iii) the goal-and-peer vector $\mathbf{g}$. The obstacle and BEV branches are encoded by compact convolutional modules, whilst $\mathbf{g}$ is processed by a small multilayer perceptron. The resulting latent features are concatenated and passed through a fusion multilayer perceptron with LayerNorm before being mapped to a Gaussian policy head producing the action mean and log-standard deviation. Actions are squashed through $\tanh$ and scaled to the bounded command space of forward velocity, yaw rate, and vertical velocity.

Training follows a CTDE formulation. During training, two centralised critics receive the concatenated per-agent latent features and joint actions, and estimate the joint action value. Target critics are maintained using Polyak averaging. During execution, each UAV acts only on its local perception and compact relative state. The SAC target is:
\begin{equation}
y = r + \gamma(1-\mathrm{done})\Big(\min_m Q^{-}_{\theta_m}(s',a') - \alpha \log \pi(a'|s')\Big)
\end{equation}
where $\alpha$ is the entropy temperature and $\mathrm{done}$ is the episode-terminal flag \cite{haarnoja2018soft}.

\subsection{Reward design}
The reward is dense, bounded, and structured to encourage goal-reaching behaviour whilst maintaining safe and smooth flight. For each agent, the reward combines progress towards the goal in the horizontal plane, reduction in vertical error, heading alignment, local clearance, action smoothness, action regularisation, and a soft penalty for violating a minimum inter-UAV separation. Terminal bonuses and penalties are applied for goal completion, collision, out-of-bounds termination, and drone-to-drone contact.

The per-agent reward is defined as:
\begin{equation}
\begin{aligned}
r_i \;=\; &\underbrace{w_{xy}\big(d_t - d_{t+1}\big)}_{\text{progress in XY}}
+ \underbrace{w_z\big(|\Delta z_t| - |\Delta z_{t+1}|\big)}_{\text{vertical progress}}
+ \underbrace{w_h \cos\theta_{t+1}}_{\text{heading}} \\
&+ \underbrace{w_c\,\mathrm{clearance}_{ego}}_{\text{navigability}} \\
&- \underbrace{w_t}_{\text{time}}
- \underbrace{w_s\|\mathbf{a}_{t+1}-\mathbf{a}_t\|^2}_{\text{smoothness}}
- \underbrace{w_v\|\mathbf{a}_{t+1}\|^2}_{\text{speed regulariser}} \\
&- \underbrace{w_{vz}^{\text{near}}\,\mathbb{I}(|\Delta z_{t+1}|<\delta)\,v_z^2}_{\text{vertical discipline}} \\
&- \underbrace{w_{vz}^{\text{wrong}}\,\mathbb{I}(|\Delta z_{t+1}|>\delta)\,
\mathbb{I}(\mathrm{sign}(v_z)=\mathrm{sign}(\Delta z))\,|v_z|}_{\text{vertical discipline}} \\
&- \underbrace{w_{\mathrm{sep}}\,\phi(d_{\text{peer}})}_{\text{separation penalty}}
\end{aligned}
\label{eq:reward}
\end{equation}
where:
\begin{equation}
\phi(d_{\text{peer}})=\left(\max\left(0,\frac{d_{\text{safe}}-d}{\sigma}\right)\right)^2
\label{eq:smooth}
\end{equation}

In Eq.~\ref{eq:reward}, $d_t$ denotes the planar distance to the goal, $\Delta z_t = z_t - z_{\text{goal}}$ is the vertical error, and $\theta_{t+1}$ is the heading error relative to the goal direction. $\mathrm{clearance}_{ego}$ denotes the local obstacle clearance, defined as the minimum distance to nearby obstacles from onboard perception (e.g., LiDAR). The action is $\mathbf{a}_t = (v, \dot{\psi}, v_z)$, where $v_z$ is the vertical velocity. The parameter $\delta$ defines a vertical tolerance band: inside the band, $v_z^2$ is penalised to stabilise altitude; outside, motion away from the target height is penalised based on $\mathrm{sign}(v_z)=\mathrm{sign}(\Delta z)$. $\mathbb{I}(\cdot)$ denotes the indicator function.

In Eq.~\ref{eq:smooth}, $\phi(d_{\text{peer}})$ is a smooth hinge penalty that is zero outside the safety radius and increases as agents approach one another. $d_{\text{peer}}$ is the inter-agent distance, with $d_{\text{safe}}$ the minimum safe separation and $\sigma$ a scaling factor. This design encourages efficient and stable navigation without rewarding unnecessary separation.

\subsection{Training procedure}
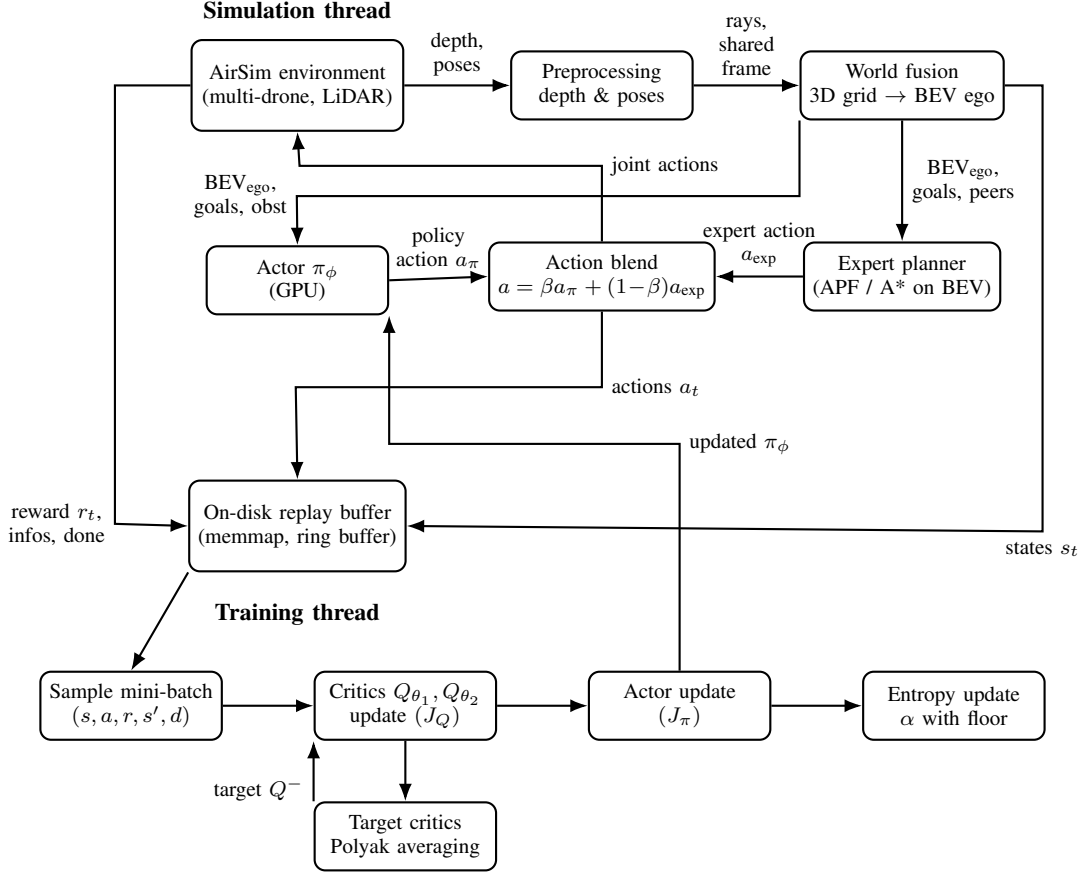
\begin{figure*}[!htbp]
\centering
\begin{tikzpicture}[
  node distance=1.0cm and 1.4cm,
  every node/.style={font=\footnotesize},
  box/.style={draw, rounded corners, thick, minimum width=2.4cm, minimum height=0.9cm, align=center},
  bigbox/.style={draw, rounded corners, thick, minimum width=2.8cm, minimum height=1.2cm, align=center},
  arr/.style={-Latex, thick}
]

\node[align=center, font=\small\bfseries] (simlabel) at (0,2.0) {Simulation thread};
\node[align=center, font=\small\bfseries] (trainlabel) at (0,-6.0) {Training thread};

\node[bigbox] (env) at (0,1) {AirSim environment\\(multi-drone, LiDAR)};
\node[box, right=of env] (pre) {Preprocessing\\depth \& poses};
\node[box, right=of pre] (fuse) {World fusion\\3D grid $\to$ BEV ego};
\node[box, below=1.6cm of fuse] (expert) {Expert planner\\(APF / A* on BEV)};
\node[box, below=1.5cm of env] (actor) {Actor $\pi_\phi$\\(GPU)};
\node[box, below=1.6cm of pre] (blend) {Action blend\\$a = \beta a_\pi + (1\!-\!\beta)a_{\text{exp}}$};

\node[bigbox, below=4.6cm of env] (rb) {On-disk replay buffer\\(memmap, ring buffer)};

\node[box, below=1.3cm of rb, xshift=-2.2cm] (batch) {Sample mini-batch\\$(s,a,r,s',d)$};
\node[box, right=1.2cm of batch] (critic) {Critics $Q_{\theta_1},Q_{\theta_2}$\\update $(J_Q)$};
\node[box, right=1.2cm of critic] (actorup) {Actor update\\$(J_\pi)$};
\node[box, right=1.2cm of actorup] (alpha) {Entropy update\\$\alpha$ with floor};

\node[box, below=0.8cm of critic] (target) {Target critics\\Polyak averaging};

\draw[arr] (env) -- node[above, align=center]{depth,\\poses} (pre);
\draw[arr] (pre) -- node[above, align=center]{rays,\\shared\\frame} (fuse);
\draw[arr] (fuse) -- node[right, align=center]{BEV$_{\text{ego}}$,\\goals, peers} (expert);
\draw[arr] (fuse.south west) -- ++(0,-1.0) |- ++(-6.65,0) node[left, align=center]{BEV$_{\text{ego}}$,\\goals, obst} -- (actor.north);
\draw[arr] (expert.west) -- node[above, align=center]{expert action\\$a_{\text{exp}}$} (blend.east);
\draw[arr] (actor.east) -- node[above, align=center]{policy\\action $a_\pi$} (blend.west);
\draw[arr] (blend.north) -- ++(0,1.0) node[right]{joint actions} |- ++(-4.0,0) --(env.south);

\draw[arr] (fuse.east) -- ++(0.5,0.0) |- ++(0,-5.9) node[below]{states $s_t$} -- (rb.east);
\draw[arr] (blend.south) -- ++(0,-1.0) node[right]{actions $a_t$} |- ++(-4.05,0) -- (rb.north);
\draw[arr] (env.west)  -- ++(-1,0) |- ++(0,-5.8) node[left, align=center]{reward $r_t$,\\infos, done} -- (rb.west);

\draw[arr] (rb.south west) -- (batch.north);
\draw[arr] (batch) -- (critic);
\draw[arr] (critic) -- (actorup);
\draw[arr] (actorup) -- (alpha);
\draw[arr] (critic.south) -- (target.north);

\draw[arr] (actorup.north) |- ++(0,3.0) node[right]{updated $\pi_\phi$}  -| (actor.south east);
\draw[arr] (target.north west) -- node[left,pos=0.2]{target $Q^{-}$} (critic.south west);

\end{tikzpicture}
\caption{Overview of the multi-agent SAC pipeline. The simulation thread runs the AirSim environment, performs depth preprocessing and world fusion, blends expert and policy actions, and writes transitions to an on-disk replay buffer. The training thread samples batches from replay to update the critics, actor, and entropy coefficient, feeding updated parameters back to the simulation thread.}
\label{fig:masac-pipeline}
\end{figure*}

Experience collection and optimisation are decoupled into separate simulation and training threads, as shown in Fig \ref{fig:masac-pipeline}. The simulation thread executes both UAVs in AirSim, performs depth preprocessing, updates the shared world map, constructs ego-aligned policy inputs, computes rewards, and appends transitions to an on-disk replay buffer. The training thread samples mini-batches from replay, updates the critics at each optimisation step, updates the actor with a delayed schedule, and applies Polyak averaging to the target networks.

To improve training stability, an expert guidance signal is used during early learning. Specifically, policy actions are blended with actions from an expert controller based on APF or A* planning over the BEV map,
\begin{equation}
a = \beta a_{\pi} + (1-\beta)a_{\text{expert}}
\end{equation}
where $a_\pi \sim \pi_\phi(\cdot \mid s)$ denotes the policy action and $a_{\text{expert}}$ is provided by a classical controller (e.g., APF or A*); the blended action is executed during data collection. The coefficient $\beta$ is gradually annealed from 0 to 1 as replay fills and policy performance improves. Additional stabilisation measures include critic and actor gradient clipping, conservative learning rates, entropy-temperature optimisation, and checkpoint selection based on policy-only success. The simulation-time training and interaction pipeline of the proposed method is summarised in Algorithm \ref{algorithm:masac}.

\begin{algorithm}[t]
\SetAlgoLined
\DontPrintSemicolon
\caption{Multi-Agent Soft Actor-Critic (MASAC) for Cooperative Drone Guidance}
\SetKwInOut{Input}{Input}
\Input{Initial policy parameters $\phi$, critic parameters $\theta_{1,2}$, target parameters $\theta_{1,2}^- \leftarrow \theta_{1,2}$, empty replay buffer $\mathcal{D}$, expert planner $\pi_{exp}$}

\For{each episode}{
    Initialize shared world voxel-map $M$ and drone states $s_i$\;
    \For{each timestep $t$}{
        \tcp{Simulation Thread: Data Collection}
        \For{each agent $i \in \{1, \dots, N\}$}{
            Extract ego-centric BEV crop $C_i$ (occupancy + clearance) from $M$\;
            Construct state $s_{i,t}$ using $C_i$, angular depth, and goal/peer vector $g_i$\;
            Sample action $a_{i,t} \sim \pi_\phi(\cdot|s_{i,t})$\;
            \tcp{Expert Action Blending}
            $a_{i,t}^{env} \leftarrow \beta a_{i,t} + (1-\beta)\pi_{exp}(s_{i,t})$\;
        }
        Execute joint actions $\mathbf{a}_t^{env}$, observe rewards $\{r_{i,t}\}$ and next states $\{s_{i,t+1}\}$\;
        Update shared map $M$ with new LiDAR observations\;
        Store transition $(\mathbf{s}_t, \mathbf{a}_t, \mathbf{r}_t, \mathbf{s}_{t+1}, \mathbf{d}_t)$ in $\mathcal{D}$\;
        
        \tcp{Training Thread: Optimisation}
        \If{it is time to update}{
            Sample minibatch $B \sim \mathcal{D}$\;
            \For{each agent $i$}{
                Compute target: $y_i = r_i + \gamma(1-d_i)(\min_{j=1,2} Q_{\theta_j}^-(\mathbf{s}', \mathbf{a}') - \alpha \log \pi_\phi(a_i'|s_i'))$\;
                Update critics by minimizing $\frac{1}{|B|}\sum (Q_{\theta_{1,2}}(\mathbf{s}, \mathbf{a}) - y_i)^2$\;
            }
            \If{$t \pmod{\text{delay}} == 0$}{
                Update actor $\phi$ using policy gradient: $\nabla_\phi J_\pi(\phi)$\;
                Update entropy coefficient $\alpha$\;
                Update target networks: $\theta_j^- \leftarrow \tau \theta_j + (1-\tau) \theta_j^-$ \;
            }
        }
        Anneal blending coefficient $\beta$ from 0 to 1 \;
    }
}
\label{algorithm:masac}
\end{algorithm}

\subsection{Real-world system integration and deployment}
A key contribution of this work is the deployment of the cooperative guidance framework on physical UAVs in GNSS-denied indoor environments. To enable this, the learned controller was embedded within a ROS~2-based onboard autonomy stack interfaced with PX4 through $\mu$XRCE. The onboard software handled LiDAR acquisition, depth-feature generation, shared-map construction, inter-UAV communication, safety monitoring, and offboard velocity command generation. Computationally demanding perception and mapping components were implemented in C++ to satisfy real-time execution requirements on the Jetson Orin Nano companion computer.

Because the real-world experiments were conducted without GNSS, vehicle state estimation was provided by DLIO LiDAR-inertial odometry \cite{chen2023direct}. The resulting pose estimate was used throughout the autonomy stack for shared-map fusion, goal representation, and closed-loop control. This provided a consistent local reference frame for both individual guidance and cooperative map sharing.

Cooperation between UAVs was supported through exchange of both pose information and batched shared-map updates. In the deployed system, positional information was transmitted at a higher rate than world-map updates, reflecting practical communication and processing constraints. Each UAV therefore combined its own sensed geometry with peer-provided map information to construct a richer spatial representation than would be available from onboard sensing alone.

\begin{figure}[!htbp]
    \centering
    \subfloat[Ego voxel map without the noise filtering.]{
        \includegraphics[width=0.4\linewidth]{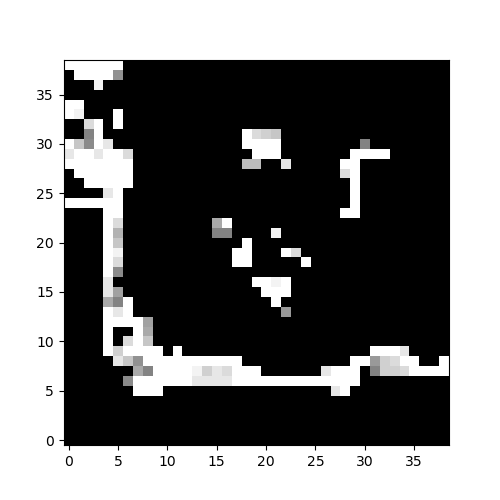}
    }
    \hfill
    \subfloat[Ego voxel map with the noise filtering.]{
        \includegraphics[width=0.4\linewidth]{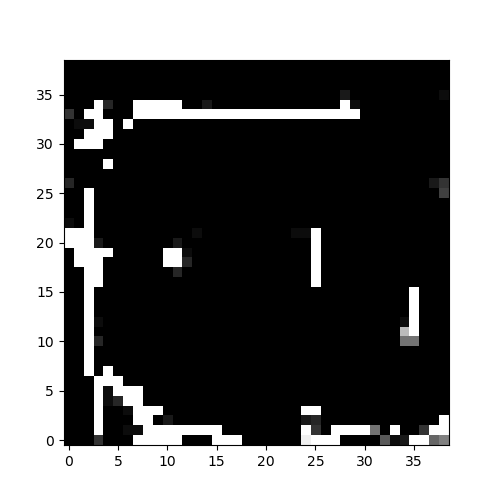}
    }
    \caption{A comparison of the ego voxel map with and without the noise filtering showing the noise around the drone that effects not only the AI guidance but also why the safety systems need a similar filtering to work.}
    \label{fig:total_figure}
\end{figure}

In real-world deployment, the voxel-based map from LiDAR observations exhibited noticeable noise, particularly around the drone body, due to sensor artefacts and structural reflections. These spurious returns can degrade both policy inputs and safety checks. To address this, a filtering stage was introduced to suppress such artefacts, as illustrated in Fig \ref{fig:total_figure}. The noise filtering results in a more stable and reliable spatial representation for both control and safety mechanisms.

To reduce deployment risk, the learned controller operated alongside conservative safety mechanisms. These included inter-UAV separation checks, LiDAR-based obstacle-proximity gating, and supervised reset procedures during staged validation. These safety layers were essential not only for protecting the hardware, but also for enabling iterative debugging and refinement of the cooperative autonomy pipeline during real-world trials.

This system integration highlights an important practical result of the study: successful transfer of cooperative DRL from simulation to hardware depends not only on the learned policy, but also on the reliability of the surrounding perception, localisation, communication, and safety infrastructure.

\subsection{Real-world adaptation}
Although the policy was trained primarily in simulation, initial hardware deployment revealed a residual sim-to-real gap caused by sensing artefacts, actuation latency, calibration mismatch, and environmental differences. To reduce this discrepancy, the simulation-trained actor was further adapted using offline imitation learning on real-world data.

During data collection, the learned policy did not directly command the platform. Instead, observations were logged while a conservative A*-based reference controller generated velocity-level actions. Each dataset sample comprised the ego-centric BEV tensor, the angular obstacle-band features, the compact goal-state vector, and the corresponding expert action. Data were collected across multiple representative arena layouts to improve environmental coverage.

The actor was then fine-tuned by behaviour cloning, minimising:
\begin{equation}
\mathcal{L}_{\mathrm{BC}}(\theta)=
\mathbb{E}_{(\mathbf{o}_t,\mathbf{a}^{\ast}_t)\sim\mathcal{D}}
\left[
\left\lVert \pi_{\theta}(\mathbf{o}_t)-\mathbf{a}^{\ast}_t \right\rVert_2^2
\right]
\end{equation}
To preserve the robust behaviour learned in simulation, fine-tuning used a small learning rate and conservative optimisation settings. In practice, the best real-world behaviour was obtained early in fine-tuning, with longer optimisation tending to degrade performance. Unless otherwise stated, the real-world results reported in this paper correspond to the best-performing fine-tuned checkpoint.

\section{Experimental Setup}

\subsection{Simulation environment and training protocol}
Training and evaluation in simulation were conducted in Unreal Engine 4 using AirSim. The simulated environment comprised a corridor-based suite designed to stress longer-horizon obstacle negotiation, constrained flight, and inter-UAV interaction. Episodes terminated upon goal completion, collision, out-of-bounds violation, or timeout.

Training was carried out on a workstation equipped with an NVIDIA RTX A4500 GPU. Approximately 100{,}000 environment steps were collected over 2{,}000 episodes. Checkpoints were retained throughout training for subsequent policy selection and evaluation.

\subsection{Model selection and baselines}
Policy selection followed a two-stage protocol. First, retained checkpoints were evaluated across all training corridors, and only policies achieving 100\% success on this suite were shortlisted. Secondly, shortlisted policies were assessed on progressively more challenging scenarios, including previously unseen corridor geometries, and the most robust policy was selected for detailed evaluation.

The proposed method was compared against three baselines:
\begin{itemize}
    \item \textbf{A* planning} operating on the world-frame BEV map,
    \item \textbf{Artificial Potential Field (APF)} control,
    \item \textbf{Sapience 1 guidance agent}, a prior policy designed primarily for short, near-linear waypoint transitions.
\end{itemize}

\subsection{Simulation evaluation}
In simulation, the selected policy was evaluated over randomly sampled corridor layouts for a fixed number of trials. A trial was counted as successful only if both UAVs reached their respective goals without collision, timeout, or out-of-bounds termination. This metric was chosen to reflect the cooperative nature of the task, since failure of either platform invalidated the joint mission.

\subsection{Real-world platform and deployment}
\begin{figure}[!htbp]
    \centering
    \includegraphics[width=0.8\linewidth]{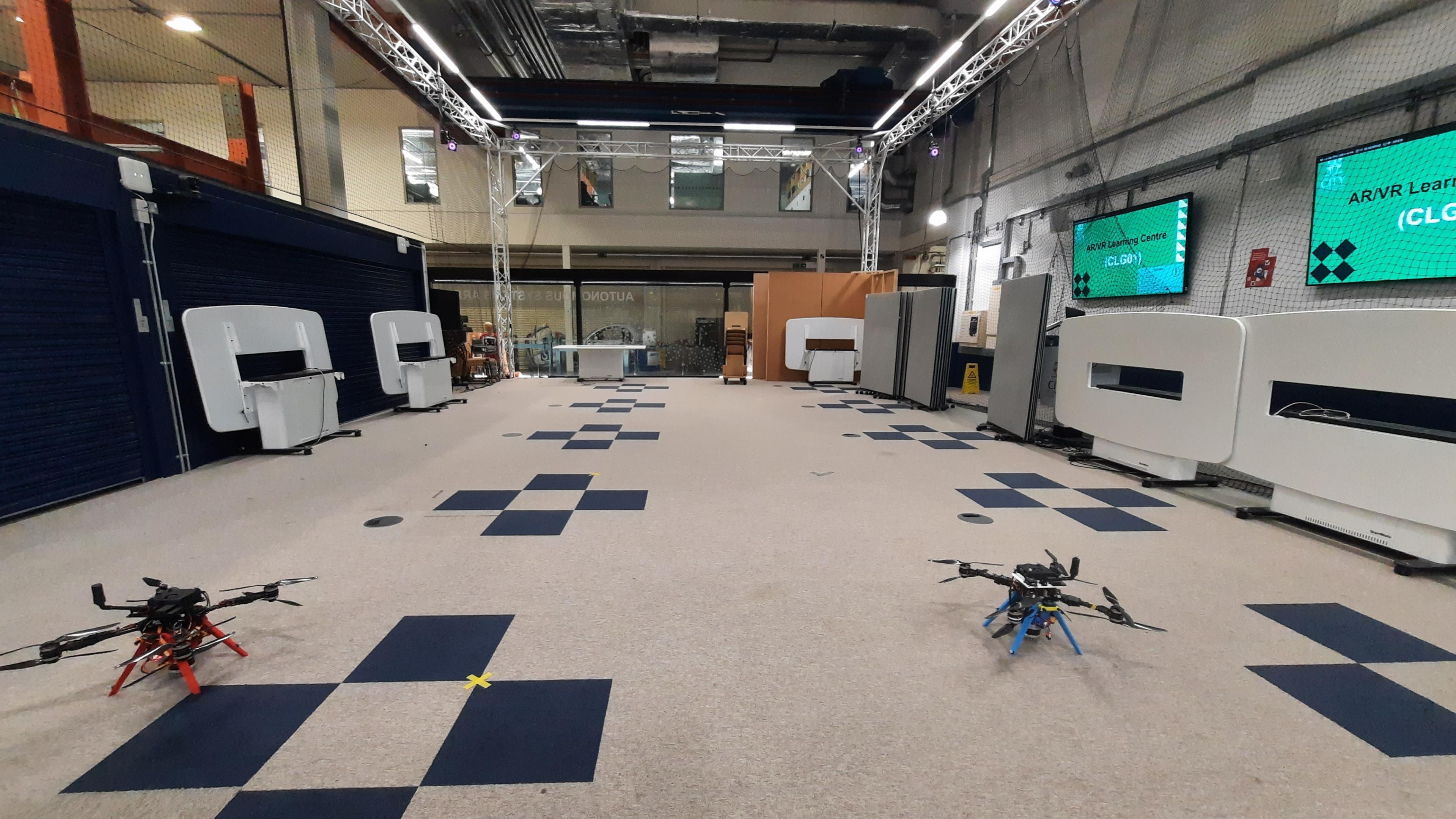}
    \caption{The City St George's University of London indoor drone flight arena.}
    \label{fig:arena}
\end{figure}

Real-world experiments were conducted in the indoor drone arena at City St George's, University of London, with approximate dimensions of $13\,\mathrm{m}\times 8\,\mathrm{m}$, as shown in Fig \ref{fig:arena}. The arena was enclosed by safety netting and configured using modular $1.2\,\mathrm{m}\times 2.4\,\mathrm{m}$ wooden wall panels to create repeatable obstacle layouts.

\begin{figure}[!htbp]
    \centering
    \includegraphics[width=1.0\linewidth]{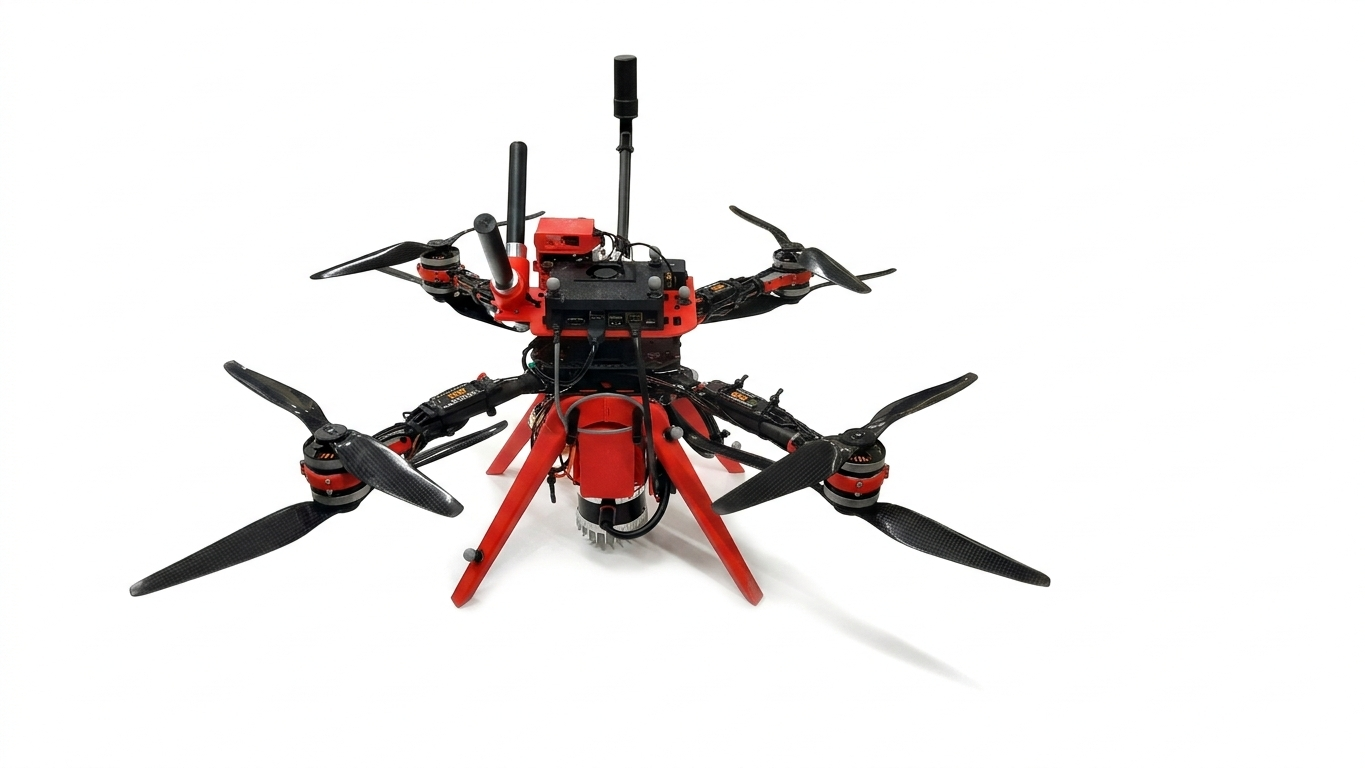}
    \caption{The octocopter drone layout with the Ouster Lidar on the bottom and the Jetson Orin Nano companion computer on top.}
    \label{fig:placeholder}
\end{figure}

The UAV platform was based on a modified Holybro X550v2 frame configured as an octocopter with contra-rotating propellers, as shown in Fig \ref{fig:placeholder}. Each vehicle used a Cube Orange+ flight controller running PX4, an NVIDIA Jetson Orin Nano companion computer, an Ouster OS-1 128-line LiDAR, and a Doodle Labs Mesh Rider radio for inter-UAV communication. Typical flight time was approximately 12 minutes using dual 5500\,mAh batteries.

The onboard autonomy stack was implemented in ROS~2, which handled sensor integration, mapping, inter-process communication, and offboard control. LiDAR-inertial odometry was provided by DLIO \cite{chen2023direct}, enabling localisation and map integration in GNSS-denied indoor flight. Conservative safety mechanisms were applied throughout deployment, including inter-UAV separation checks, obstacle-proximity gating, and supervised reset procedures during early validation.

\subsection{Real-world evaluation scenarios}
Real-world testing proceeded incrementally. Initial single-UAV experiments were used to validate perception, mapping, odometry, and safety logic. Two-UAV trials were then introduced to verify shared-map fusion and cooperative operation before progressing to obstacle-rich scenarios.

\begin{figure}[!htbp]
    \centering
    \includegraphics[width=0.8\linewidth]{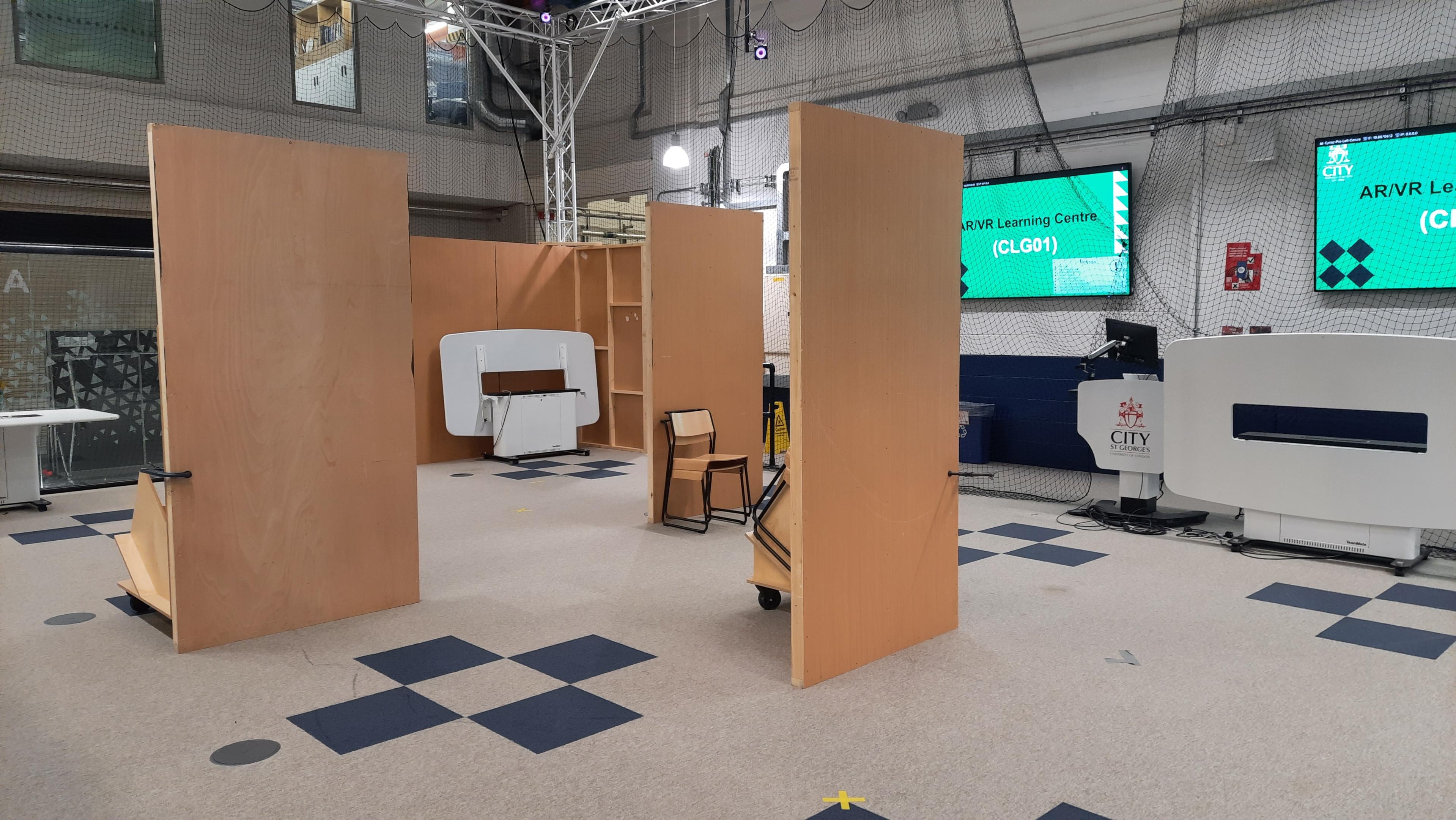}
    \caption{The three-panel course in the arena.}
    \label{fig:3panel}
\end{figure}

Five arena configurations were evaluated: i) an empty arena, ii) a single-panel obstacle, iii) a two-panel layout, iv) a three-panel layout, as shown in Fig \ref{fig:3panel}, and v) a T-shaped course. A run was counted as successful only if the task was completed without collision with obstacles or the peer UAV, without safety-triggered termination, and without manual intervention.

\section{Results and Discussion}

\subsection{Simulation performance}
The selected MASAC policy achieved a 90.3\% success rate across the simulated corridor evaluation suite (Table~\ref{tab:success_rates}), outperforming all baseline methods. The strongest baseline was A*, which achieved 55.0\% success, but its performance was limited by delayed or noisy map updates, infrequent replanning, and conservative treatment of the peer UAV as an obstacle within the shared map. Under these conditions, A* could stall, generate unnecessarily long detours, or produce intersecting trajectories when peer-state updates were imperfect. The APF controller performed substantially worse, particularly in dead-ends and maze-like layouts, where purely local repulsion and attraction cues provided insufficient look-ahead. The prior Sapience~1 guidance agent also underperformed, reflecting the fact that it was designed primarily for short, near-linear waypoint transitions rather than cooperative long-horizon navigation.

\begin{table}[!htbp]
\centering
\rowcolors{2}{gray!30}{gray!10}
\caption{Success rate across the corridor evaluation suite.}
\label{tab:success_rates}
\begin{tabular}{|l|c|p{3.4cm}|}
\hline
\textbf{Method} & \textbf{Success rate} & \textbf{Notes} \\
\hline
MASAC agent & \textbf{90.3\%} & Learned multi-input controller using depth, BEV, and relative state \\
A* planner & 55.0\% & Limited by replanning and treatment of peer UAVs as obstacles \\
APF controller & 21.5\% & Local-only; prone to dead-ends and oscillation \\
Sapience~1 agent & 30.0\% & Designed for short waypoint transitions rather than cooperative navigation \\
\hline
\end{tabular}
\end{table}

\begin{figure}[!htbp]
    \centering
    \includegraphics[width=\linewidth]{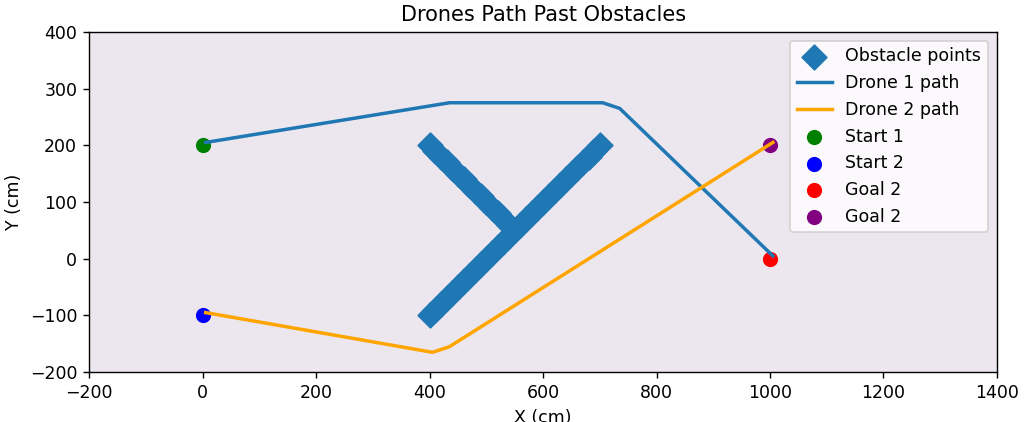}
    \caption{Representative trajectory of the selected agent through one of the evaluation corridors.}
    \label{fig:drone_path}
\end{figure}

These results indicate that the combination of shared world-frame fusion, ego-aligned BEV control input, and dense reward shaping provides a more robust basis for cooperative indoor navigation than either purely reactive control or classical planning under practical sensing and multi-agent constraints. A representative simulated trajectory is shown in Fig.~\ref{fig:drone_path}.

\subsection{Real-world performance}
Real-world evaluation was conducted after offline imitation-based fine-tuning using state-action pairs labelled by an A*-based reference controller. This adaptation stage was necessary because the simulation-trained policy alone exhibited reduced stability under real sensing and actuation conditions. Following fine-tuning, the controller was evaluated in two-UAV operation across five indoor arena configurations of increasing difficulty.

A practical constraint during deployment was the shared-map update rate, which operated at approximately 5\,Hz, with higher-rate positional exchange between UAVs. During early trials, a defect in the map-fusion pipeline prevented occupancy decay and led to saturation of the shared map over time. After separating the combined and uncombined map streams and restoring temporal decay, the mapping process remained numerically stable and the experimental campaign proceeded successfully.

Table~\ref{tab:real_world_success} summarises the real-world outcomes. Across all five arena configurations, the system achieved 100\% success over 10 trials per course, with both UAVs completing the task without collision, manual intervention, or safety-triggered termination.

\begin{table}[!htbp]
\centering
\rowcolors{2}{gray!30}{gray!10}
\caption{Real-world performance across five indoor arena configurations. Each course was executed for 10 trials with two UAVs. A success is counted when both UAVs complete the objective without collision or manual intervention.}
\label{tab:real_world_success}
\begin{tabular}{|l|c|c|}
\hline
\textbf{Course} & \textbf{Trials} & \textbf{Success rate} \\
\hline
Empty arena & 10 & 100\% \\
Single-panel obstacle & 10 & 100\% \\
Two-panel obstacle & 10 & 100\% \\
Three-panel obstacle & 10 & 100\% \\
T-shaped layout & 10 & 100\% \\
\hline
\end{tabular}
\end{table}

\begin{figure}[!htbp]
    \centering
    \includegraphics[width=1.0\linewidth]{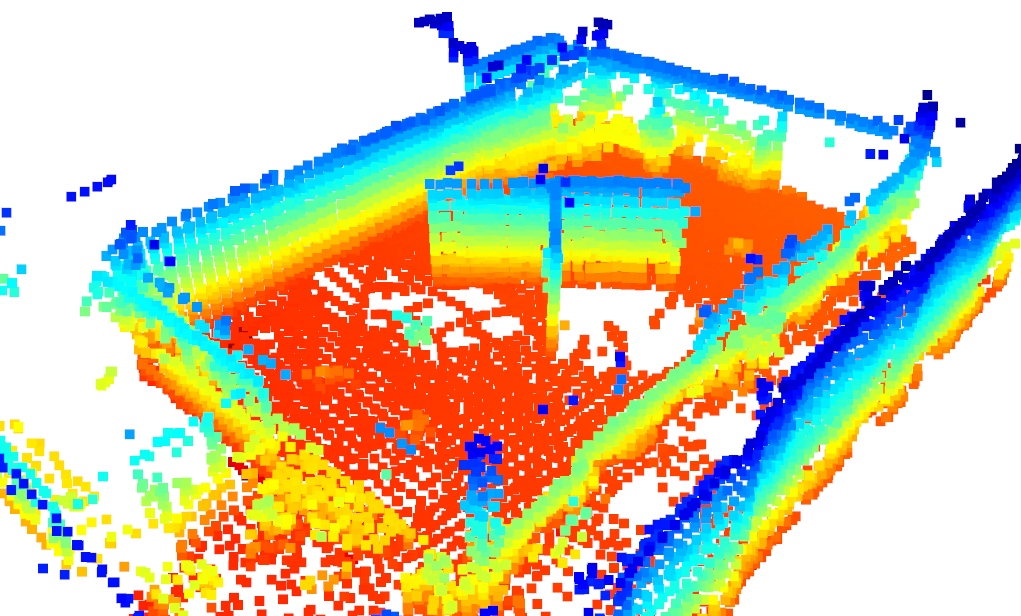}
    \caption{The combined feature maps from the DLIO packages on both drones.}
    \label{fig:pointcloudtshaped}
\end{figure}

\begin{figure}[!htbp]
    \centering
    \includegraphics[width=1.0\linewidth]{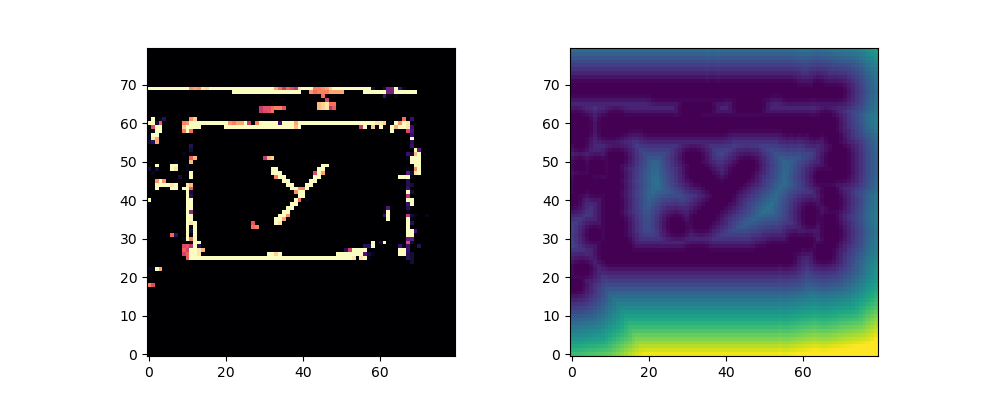}
    \caption{The benefits of the voxel map sharing can be seen in this image as the entire wall structure is depicted in the voxel map even though it can not all be seen by one drone.}
    \label{fig:voxelmapandclearence}
\end{figure}

In Fig.~\ref{fig:pointcloudtshaped}, combined DLIO-derived maps from both UAVs showed that shared perception improved reconstruction of the obstacle layouts beyond what either individual platform could observe alone. As shown in Fig.~\ref{fig:voxelmapandclearence}, this was particularly evident in the more constrained courses, where the shared map provided mid-field context that supported route selection before the full obstacle structure was directly visible to a single vehicle.

\subsection{Qualitative behavioural observations}

\begin{figure}[!htbp]
    \centering
    \includegraphics[width=1.0\linewidth]{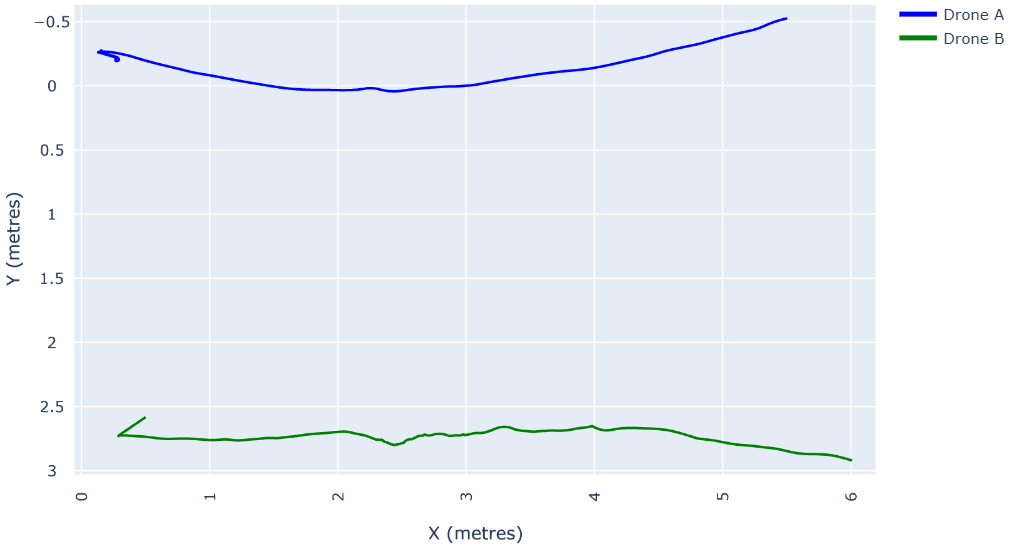}
    \caption{The top-down view of the initial test of the multi-drone deployment in the open arena with no obstacles. The drones move towards the target points in an efficient manner.}
    \label{fig:2d_nopanel}
\end{figure}

\begin{figure}[!htbp]
    \centering
    \includegraphics[width=1.0\linewidth]{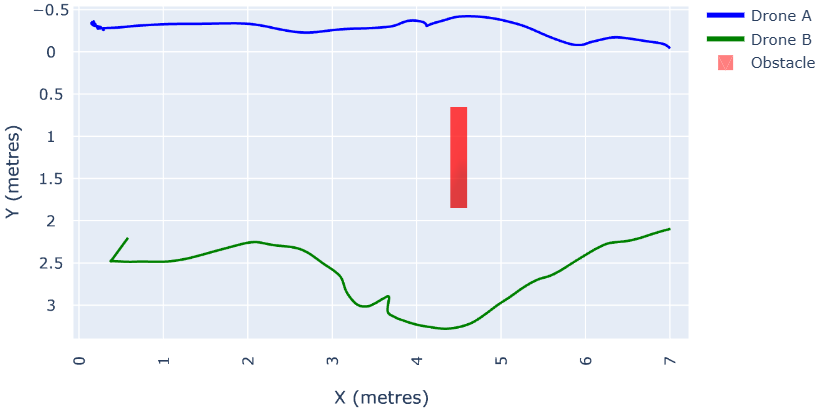}
    \caption{The top-down view of the one-panel obstacle flight. Some obstacle avoidance behaviour is shown.}
    \label{fig:2d_1panel}
\end{figure}

\begin{figure}[!htbp]
    \centering
    \includegraphics[width=1.0\linewidth]{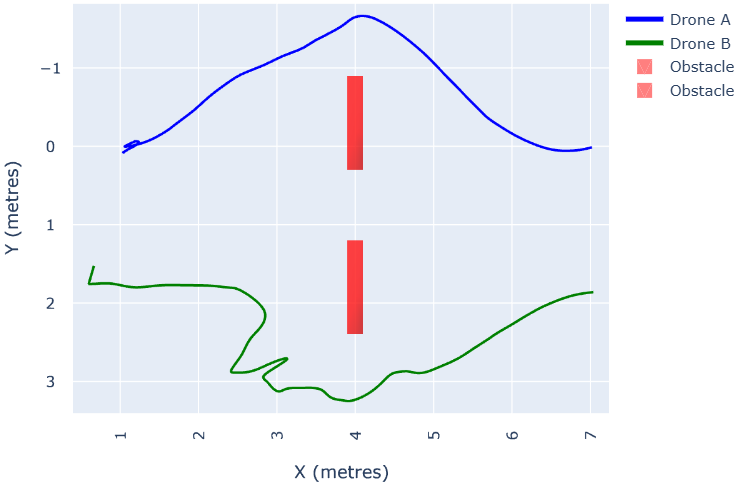}
    \caption{The top-down view of the two obstacle flight. The gap between the panels is not large enough to fly a drone through safely, so the drones avoid it. The path shows the safety measure built into the software, pushing Drone B wider due to an abundance of caution.}
    \label{fig:2d_2panel}
\end{figure}

Although all real-world courses were completed successfully, the different layouts exposed distinct aspects of the controller behaviour. In the empty arena, both UAVs reached their goals efficiently whilst maintaining stable altitude, confirming nominal goal-seeking behaviour in the absence of obstacles as shown in Fig.~\ref{fig:2d_nopanel}. In the single-panel and two-panel layouts, both vehicles consistently selected feasible routes around obstacles rather than attempting to force passage through geometrically unsafe gaps, as shown in Fig.~\ref{fig:2d_1panel} and Fig.~\ref{fig:2d_2panel}. In particular, the two-panel course showed that the controller did not simply pursue the shortest geometric route, but instead favoured wider and safer trajectories when the central gap was below the practical traversal width of the platform.

\begin{figure}[!htbp]
    \centering
    \includegraphics[width=1.0\linewidth]{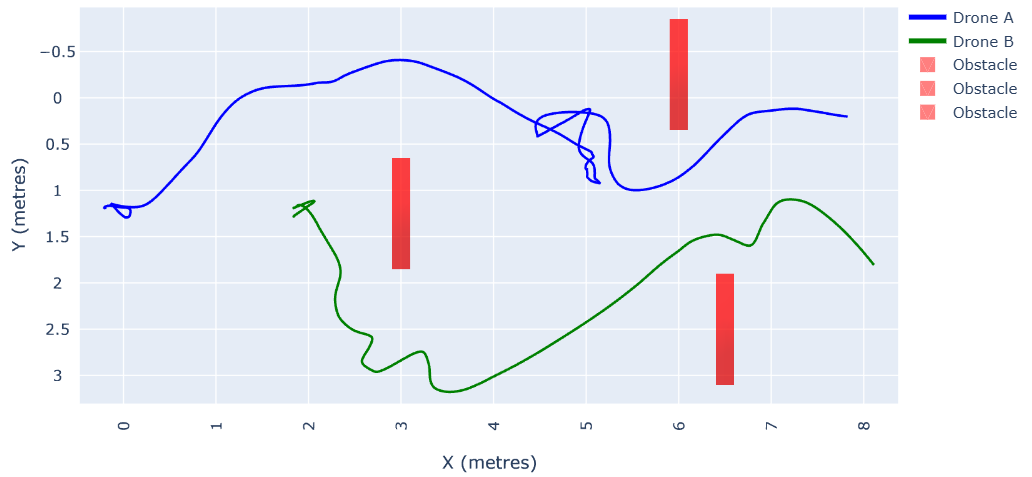}
    \caption{The top-down view of the flight path of the three-panel obstacle course. Drone A can be seen to hold while Drone B is navigating the gap.}
    \label{fig:2d_3panel}
\end{figure}

\begin{figure}[!htbp]
    \centering
    \includegraphics[width=1.0\linewidth]{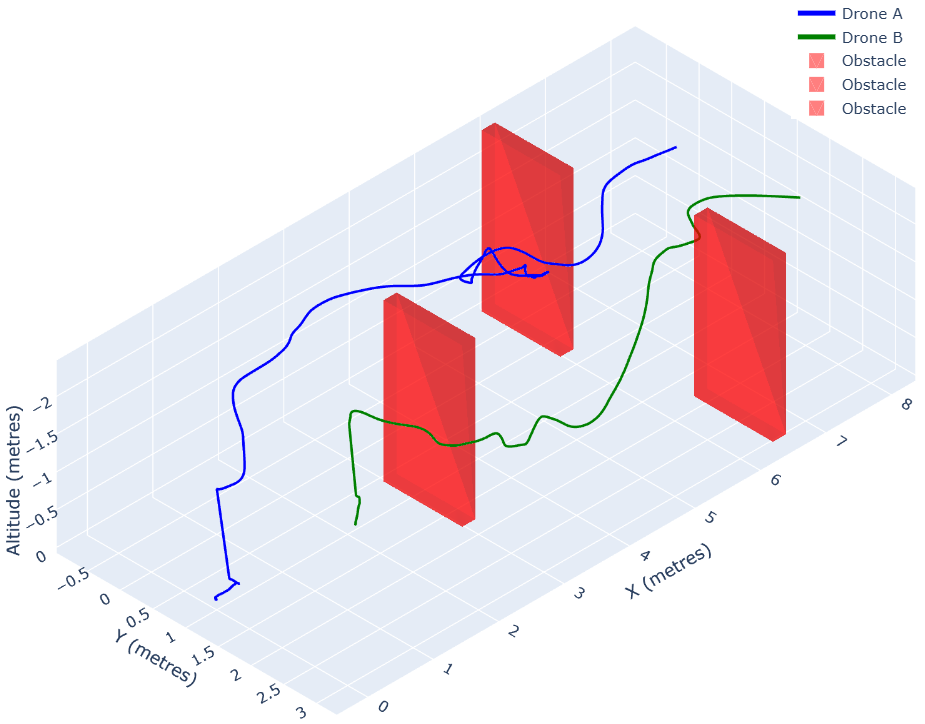}
    \caption{The isometric view of the three-panel obstacle course with the two drones path through it.}
    \label{fig:3d_3panel}
\end{figure}

The three-panel layout provided the clearest evidence of cooperative interaction in constrained geometry. In this scenario, one UAV temporarily held position while the other cleared the narrow passage, after which it resumed progress towards its own goal, as shown in Fig.~\ref{fig:2d_3panel} and Fig.~\ref{fig:3d_3panel}. This behaviour was not imposed by an explicit sequencing module, but emerged from the combination of shared-map fusion, inter-UAV separation logic, and learned continuous control. 

\begin{figure}[!htbp]
    \centering
    \includegraphics[width=1.0\linewidth]{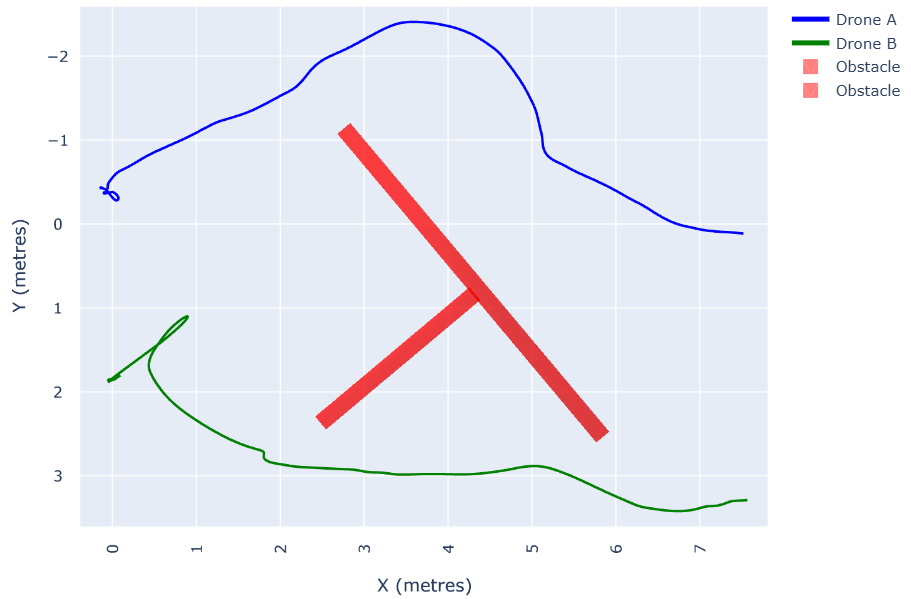}
    \caption{Top-down view of the two drones flying around the T-shaped obstacle used in previous studies.}
    \label{fig:2dtshaped}
\end{figure}

In the T-shaped layout, the two UAVs selected opposite routes around the obstacle structure, again consistent with cooperative operation under shared perception, as shown in Fig.~\ref{fig:2dtshaped}.

\subsection{Deployment lessons and limitations}
The real-world experiments highlight several lessons that are less visible in simulation. First, modest offline adaptation using real-world state-action data was important for stable transfer from simulation to hardware. Secondly, shared-map fusion must be engineered carefully: communication-limited update rates and numerical artefacts in occupancy accumulation can materially affect behaviour if not handled correctly. Thirdly, conservative safety gating improved robustness during deployment, but could also induce avoidable detours when thresholds were set aggressively.

The results demonstrate that the proposed framework can support reliable cooperative indoor UAV guidance in GNSS-denied environments, but several limitations remain. The current study considers two UAVs and a predominantly planar BEV representation. Scaling to larger teams and more vertically complex environments will likely require more efficient communication, richer 3D spatial representations, and tighter integration between learned control and deployment-time safety mechanisms. In addition, take-off transients remained a practical challenge in some runs, indicating a need for more explicit handling of early-flight stabilisation in future work.

\section{Conclusion}
This paper presented a cooperative indoor UAV guidance framework that combines a shared voxel-map world model with a multi-agent Soft Actor-Critic controller. Multiple UAVs fuse LiDAR observations into a shared world-frame representation, from which each agent receives an ego-aligned bird’s-eye-view crop containing occupancy and clearance information for decentralised continuous control. This design couples explicit geometric structure with learned cooperative guidance, providing a scalable spatial substrate for multi-UAV navigation in cluttered indoor environments.

In simulation, the proposed method outperformed classical baselines, demonstrating that shared map-based intermediate representations can support more robust cooperative navigation than purely reactive control or conventional planning under practical sensing and multi-agent constraints. Real-world experiments in a GNSS-denied indoor arena further showed that, after modest offline imitation-based adaptation, the proposed framework could support stable two UAV operation across progressively more challenging obstacle layouts. The results also highlighted the importance of system-level factors such as shared-map stability, communication constraints, and safety-gated deployment when transferring learned cooperative control from simulation to hardware. The results highlight that successful cooperative DRL deployment depends not only on policy learning, but also on the fidelity and numerical stability of the shared representation used at run time.

Overall, the study shows that shared voxel/BEV mapping combined with decentralised learned control is a practical and effective approach for cooperative indoor UAV guidance. Future work will focus on scaling to larger teams, extending the representation to more fully 3D environments, and improving robustness to deployment time effects such as take-off transients, localisation drift, and map update latency.

\bibliographystyle{IEEEtran}
\bibliography{refs}

\end{document}